\documentclass[10pt, a4paper]{article}
\usepackage{lrec}
\usepackage{graphicx}
\usepackage{tabularx}
\usepackage{soul}
\usepackage{times}
\usepackage{latexsym}
\usepackage{booktabs}
\usepackage{graphicx}
\usepackage{pgfplots}
\usepackage{floatrow}

\usepackage{caption}
\usepackage{subcaption}
\usepackage{floatrow}

\usepackage{times}
\usepackage{enumitem}
\usepackage{latexsym}
\usepackage{booktabs}
\usepackage{ltxtable} 
\usepackage{tabularx}
\usepackage{amsmath}
\usepackage{soul}

\usepackage{array, makecell}
\usepackage{boldline}

\usepackage{epstopdf}
\usepackage[latin1]{inputenc}

\usepackage{hyperref}
\usepackage{xstring}

\title{Abstractive Document Summarization without Parallel Data}

\name{Nikola I. Nikolov {\normalfont and} Richard H.R. Hahnloser}

\address{Institute of Neuroinformatics, University of Z{\"u}rich and ETH Z{\"u}rich, Switzerland \\
  {\tt \{niniko, rich\}@ini.ethz.ch}}

\abstract{
    Abstractive summarization typically relies on large collections of paired articles and summaries. However, in many cases, parallel data is scarce and costly to obtain. We develop an abstractive summarization system that relies only on large collections of example summaries and non-matching articles. Our approach consists of an unsupervised sentence extractor that selects salient sentences to include in the final summary, as well as a sentence abstractor that is trained on pseudo-parallel and synthetic data, that paraphrases each of the extracted sentences. We perform an extensive evaluation of our method: on the CNN/DailyMail benchmark, on which we compare our approach to fully supervised baselines, as well as on the novel task of automatically generating a press release from a scientific journal article, which is well suited for our system. We show promising performance on both tasks, without relying on any article-summary pairs.
 \\ \newline \Keywords{automatic summarization, abstractive summarization, text rewriting, low-resource settings} }

\begin{document}

\maketitleabstract

\section{Introduction}

Text summarization aims to produce a shorter, informative version of an input text. While extractive summarization only selects important sentences from the input, abstractive summarization generates content without explicitly re-using whole sentences \cite{nenkova2011automatic} resulting summaries that are more fluent. In recent years, a number of successful approaches have been proposed for both extractive \cite{nallapati2017summarunner,refresh} and abstractive \cite{see2017get,chen2018fast} summarization paradigms. State-of-the-art abstractive approaches are supervised, relying on large collections of paired articles and summaries. However, competitive performance of abstractive systems remains a challenge when the availability of parallel data is limited, such as in low-resource domains or for languages other than English. 

\paragraph{}

Even when parallel data is severely limited, we may have access to example summaries and large collections of articles on similar topics. Examples are blog posts or scientific press releases, for which the original articles may be unavailable or behind a paywall. 

\paragraph{}

\begin{figure}
    \centering
    \includegraphics[width=0.8\textwidth]{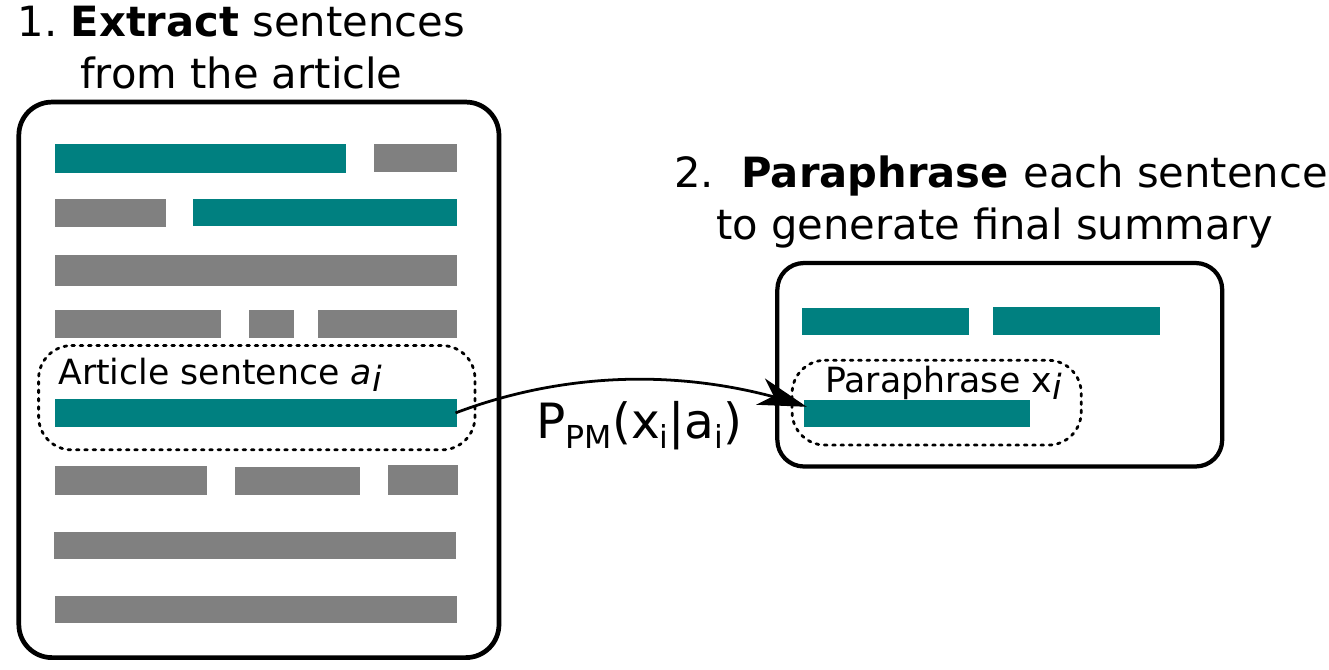}
    \caption{Overview of our approach to abstractive document summarization without parallel data: given an input article $\pmb{a}={a_1,...,a_N}$ consisting of $N$ sentences, (1) we fist select salient sentences $a_i$ (in blue) using an unsupervised extractive summarization algorithm; we then (2) generate paraphrases $x_i$ using a sentence paraphrasing model $P_{PM}$, trained on pseudo-parallel and synthetic data.}
    \label{fig:approach}
\end{figure}

In this paper, we develop a system\footnotemark\footnotetext{\url{https://github.com/ninikolov/low_resource_summarization}} for abstractive document summarization (Figure \ref{fig:approach}) that only relies on example summaries and non-matching articles, bypassing the need for large-scale parallel corpora. Our system consists of two components: First, an unsupervised \textbf{sentence extractor} selects salient sentences. Second, each extracted sentence is paraphrased using a \textbf{sentence abstractor}. The abstractor is trained on pseudo-parallel data extracted from raw corpora, as well as on additional synthetic data generated through backtranslation. 

\paragraph{}

We evaluate our approach on two summarization tasks. First, on the CNN/DailyMail news article summarization dataset, we compare the performance of our method based on non-parallel data to fully supervised models based on parallel data (Section \ref{sec:automatic-eval}). Second, we test our system on the novel task of automatically generating a press release from a scientific journal article (Section \ref{sec:science}). This task is well suited for our system because only a small number of aligned document pairs can be extracted from available repositories of scientific journal articles and press releases. On both tasks, our method achieves promising results without relying on any parallel article-summary pairs. 

\section{Background}

\subsection{Supervised Summarization}

In recent years, there have been large advances in supervised \textbf{abstractive summarization}, for headline generation \cite{rushneural,nallapati2017summarunner} as well as for generation of multi-sentence summaries \cite{see2017get}. State-of-the-art approaches are typically trained to generate summaries either in a fully end-to-end fashion \cite{see2017get}, processing the entire article at once, or hierarchically, first extracting content and then paraphrasing it sentence-by-sentence \cite{chen2018fast}. 

\paragraph{}

Both approaches rely on large collections of article-summary pairs such as the annotated Gigaword \cite{napoles2012annotated} or the CNN/DailyMail dataset \cite{nallapati2016sequence}. The heavy reliance on manually curated resources prohibits the use of abstractive summarization in domains other than news articles, or languages other than English, where parallel data may not be as abundantly available. In such areas, extractive summarization often remains the preferred choice. 

\begin{figure*}
    \centering
    \includegraphics[width=0.8\textwidth]{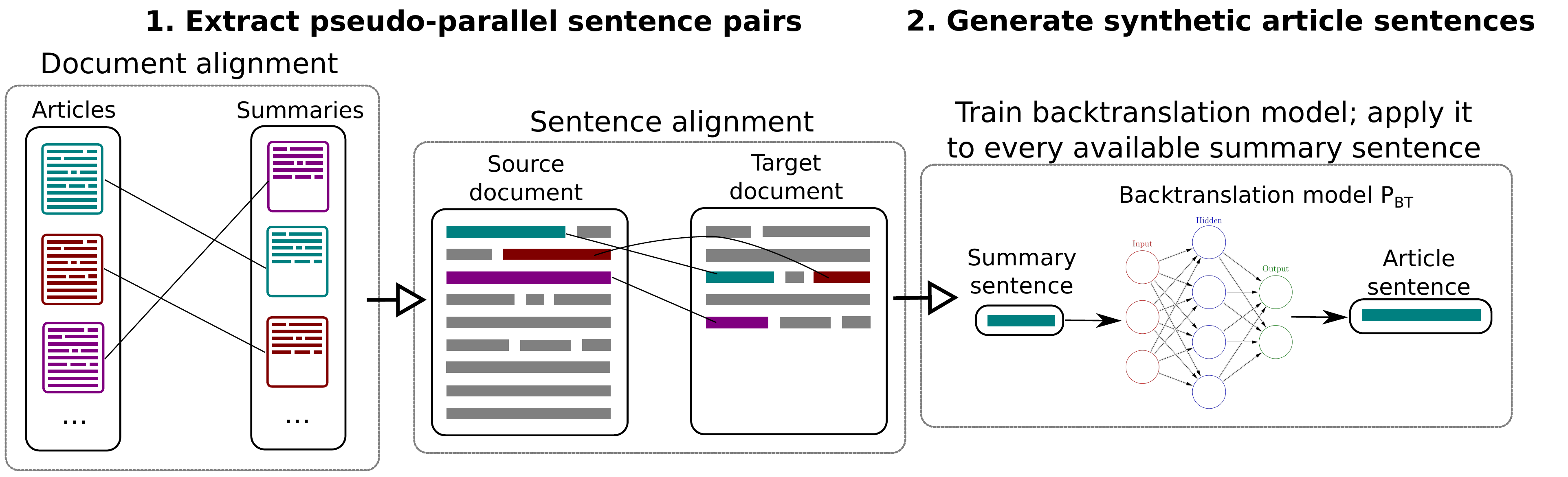}
    \caption{Overview of our pipeline for constructing the training dataset of our sentence abstractor $P_{PM}$. (1), given a dataset of summaries and articles on similar topics, we extract pseudo-parallel document and sentence pairs using large-scale alignment. (2), we use the pseudo-parallel pairs to train a backtranslation model $P_{BT}$, which we use to synthesize article sentences given a summary sentence.}
    \label{fig:paraphrasing_pipeline}
\end{figure*}

\subsection{Unsupervised Summarization}

Unsupervised summarization has a long history within the extractive summarization paradigm. Given an input article consisting of $N$ sentences $\pmb{a} = \{a_1,...,a_{N}\}$, the goal of \textbf{extractive summarization} is to select the $K$ most salient sentences as the output summary, without employing any paraphrasing or fusion. A typical approach is to weigh each sentence either with respect to the document as a whole \cite{radev2004centroid} or through an adjacency-based measure of sentence importance \cite{erkan2004lexrank}.

\paragraph{}

There is less work on unsupervised abstractive summarization. 
 \cite{chu2019meansum} propose a review summarization system based on autoencoders. Their focus is, however, on multi-document rather than on single-document summarization as it is in our case.
\cite{dohare2018unsupervised} develop a pipeline for semantic abstractive summarization which works by constructing a graph from the article and then generating a summary from the most informative part of the graph. \cite{isonuma-etal-2019-unsupervised} propose an abstractive summarization framework based on learning the discourse structure of input articles, and generating a single-sentence summary using a language model trained to reconstruct sentences from example reviews. 

\paragraph{}

Our work focuses on multi-sentence abstractive summarization using large-scale non-parallel resources such as collections of summaries \textit{without} matching articles. Recently, several methods have been proposed to reduce the need for parallel data either through harvesting pseudo-parallel data from raw corpora \cite{nikolov2018large} or by synthesizing data using backtranslation \cite{sennrich2015improving}. Such methods have been shown to be viable for a number of tasks such as unsupervised machine translation \cite{lample2018phrase}, sentence compression \cite{compression}, and style transfer \cite{lample2018multipleattribute}. To the best of our knowledge, our work is the first to extend such methods to single-document summarization to generate multi-sentence abstractive summaries in a data-driven fashion. 

\section{Approach}\label{sec:approach}

Our system (see Figure \ref{fig:approach}) consists of two components: an \textbf{extractor} (Section \ref{sec:extractor}) that picks salient sentences to include in the final summary and an \textbf{abstractor} (Section \ref{sec:abstractor}) that subsequently paraphrases each of the extracted sentences, rewriting them to meet the target summary style. 

\paragraph{}

Our approach is similar to \cite{chen2018fast}, except that they use parallel data to train their extractors and abstractors. In contrast, during training, we only assume access to $M$ example summaries $\pmb{S} = \{\pmb{s}_0,..,\pmb{s}_M\}$ without matching articles. During testing, given an input article consisting of $N$ sentences $\pmb{a} = \{a_0,..., a_N\}$, our system is capable of generating a multi-sentence abstractive summary consisting of $K$ sentences (where $K$ is a hyperparameter).

\subsection{Sentence Extractor}\label{sec:extractor}

The extractor selects the $K$ most salient article sentences to include in the summary. We consider two unsupervised variants for the extractor:

\paragraph{{\sc Lead}} picks the first \textit{K} sentences from the article and returns them as the summary. For many datasets, such as CNN/DailyMail, {\sc Lead} is a simple but tough baseline to beat, especially using abstractive methods \cite{see2017get}. Because {\sc Lead} may not be the optimal choice for other domains or datasets, we experiment with another unsupervised extractive approach. 

\paragraph{{\sc LexRank}} \cite{erkan2004lexrank} represents the input as a highly connected graph in which vertices represent sentences and edges between sentences are assigned weights equal to their term-frequency inverse document frequency (TF-IDF) similarity, provided that the TF-IDF similarity is higher than a predefined threshold $t$. The centrality of a sentence is then computed using the PageRank algorithm. 

\subsection{Sentence Abstractor}\label{sec:abstractor}

The sentence abstractor ($P_\text{PM}$) is trained to generate a paraphrase $x_i$ for every article sentence $a_i$, rewriting it to meet the target sentence style of the summaries. We implement $P_\text{PM}$ as an LSTM encoder-decoder with an attention mechanism \cite{bahdanau2014neural}. Instead of training the abstractor on parallel examples of sentences from articles and summaries, we train it on a synthetic dataset created in two steps (summarized in Figure \ref{fig:paraphrasing_pipeline}): 

\paragraph{Pseudo-parallel dataset.} The first step is to obtain an initial set of pseudo-parallel article-summary sentence pairs. Because we assume access to \textit{some} example summaries, our approach is to align summary sentences to an external corpus of comparable articles. Here, we apply the large-scale alignment method from \cite{nikolov2018large}, which hierarchically aligns documents followed by sentences in the two datasets (see Figure \ref{fig:paraphrasing_pipeline}, step 1). The alignment is implemented using nearest neighbor search: first on document and then on sentence embeddings. 

\paragraph{Backtranslated pairs.} We use the initial pseudo-parallel dataset to train a backtranslation model $P_\text{BT}(x_i|s_i)$, following \cite{sennrich2015improving}. The model learns to synthesize "fake" article sentences given a summary sentence (see Figure \ref{fig:paraphrasing_pipeline}, 2). We use $P_\text{BT}$ to generate multiple synthetic article sentences $x_{ij}$ for each summary sentence $s_i$ available, taking the $j=1,\dots,J$ top hypotheses predicted by beam search\footnote{We also experimented with sampling \cite{edunov2018understanding} but found it to be too noisy in the current setting.}. 

\paragraph{}

To train our final sentence paraphrasing model $P_\text{PM}(s_i|x_i)$, we combine all pseudo-parallel and backtranslated pairs into a single dataset of article-summary sentence pairs. 

\section{Experimental Set-up}

\begin{table*}
\RawFloats
	\begin{minipage}{0.54\linewidth}
		\centering
		\small 
\begin{tabular}{c|c|c|c|c|c}
\hlineB{2}
\textbf{Approach} & \textbf{R-1} & \textbf{R-2} & \textbf{R-L} & \textbf{MET} & \textbf{\#} \\ \hlineB{2}
{\sc Oracle} & 47.33 & 26.43 & 43.69 & 30.76 & 132 \\ \hlineB{2}
\multicolumn{6}{c}{\textbf{Unsupervised extractive baselines}} \\ \hlineB{2}
{\sc Lead} & 38.78 & 17.57 & 35.49 & 23.67 & 119 \\ 
{\sc LexRank} & 34.49 & 14.1 & 31.32 & 21.27 & 133 \\ \hlineB{2}
\multicolumn{6}{c}{\textbf{Supervised abstractive baselines (Trained on parallel data)}} \\ \hlineB{2}
{\sc LSTM} & 35.61 & 15.04 & 32.7 & 16.24 & 58 \\ 
{\sc Ext-Abs$^\dagger$} & 38.38 & 16.12 & 36.04 & 19.39 &  \\ 
{\sc Ext-Abs-RL$^\dagger$} & 40.88 & 17.8 & 38.54 & 20.38 & 73 \\ \hlineB{2}
\multicolumn{6}{c}{\textbf{Abstractive summarization without parallel data (this work)}} \\ \hlineB{2}
{\sc Lead + Abs\textsubscript{pp+syn-5}} & 32.98 & 11.13 & 30.88 & 13.51 & 50 \\ 
{\sc LexRank + Abs\textsubscript{pp+syn-5}} & 30.87  & 9.42 & 28.82 & 12.51 & 52 \\ \hlineB{2}
\end{tabular}
\protect\caption[position=bottom]{Metric results on the CNN/Daily Mail test set. \textbf{R-1/2/L} are the \textbf{ROUGE-1/2/L} F1 scores; \textbf{MET} is \textbf{METEOR}, while \textbf{\#} is the average number of tokens in the summaries. $\dagger$ are from \cite{chen2018fast}.}
\label{tab:results-cnndm}
	\end{minipage}\hfill
	\begin{minipage}{0.44\linewidth}
		\centering
		\small 
		\begin{tabular}{p{7.2cm}}
    \hlineB{2} 
    { (1) \sc Abs\textsubscript{pp-0.63}:} cnn is the first time in three years . the other contestants told the price of the price . the game show will be hosted by the tv game show . the game of the game is the first of the show . \\ \hline
     { (2) \sc Abs\textsubscript{pp+syn-5}:}
      a tv legend has returned to the first time in eight years .  contestants told the price of `` the price is right ''  bob barker hosted the tv game show for 35 years .  the game is the first of the show 's `` lucky seven '' \\ \hline
    { (3) \sc Abs\textsubscript{par}:} a tv legend returned to doing what he does best . contestants told to `` come on down ! '' on april 1 edition . he hosted the tv game show for 35 years before stepping down in 2007 . barker handled the first price-guessing game of the show , the classic `` lucky seven ''
     \\ \hlineB{2}
    \end{tabular}
	\caption{Example outputs on the CNN/DM dataset ({\sc Lead} extractor): (1)/(2) are trained on pseudo-parallel/synthetic data, while the abstractor in (3) is trained on parallel data.}
	\label{tab:examples}
	\end{minipage}
\end{table*}
\begin{table}[]
    \centering
    \small
    \begin{tabular}{c|c|c|c|c}
    \hlineB{2}
    \textbf{Approach (\# pairs)} & \textbf{R-1} & \textbf{R-2} & \textbf{R-L} & \textbf{\#} \\ \hlineB{2}
    {\sc Abs\textsubscript{pp-0.60}} (2M) & 23.08 & 4.06 & 21.48 & 62 \\ \hline
    {\sc Abs\textsubscript{pp-0.63}} (1.2M) & 28.08 & 7.07 & 26.14 & 49  \\ \hline
    {\sc Abs\textsubscript{pp-0.67}} (0.3M) & 24.36 & 4.76 & 22.64 & 57  \\ \hlineB{2}
    {\sc Abs\textsubscript{pp+syn-1}} (2.4M) & 31.92 & 10.2 & 29.9 & 51  \\ \hline
    {\sc Abs\textsubscript{pp+syn-5}} (6.6M) & 32.98 & 11.13 & 30.88 & 50 \\ \hline
    {\sc Abs\textsubscript{pp+syn-10}} (12M) & 32.8 & 11.2 & 30.71 & 49 \\  \hlineB{2}
    {\sc Abs\textsubscript{pp-ub}} (575K) & 38.42 & 15.98 & 35.8 & 65 \\ \hline
    {\sc Abs\textsubscript{par}} (1M) & 38.68 & 16.36 & 36.15 & 62 \\ \hlineB{2}
    \end{tabular}
    \caption{Comparison of abstractors trained on parallel ({\sc Abs\textsubscript{par}}) vs. pseudo-parallel data ({\sc Abs\textsubscript{pp-$\theta_s$}}, using different sentence alignment thresholds $\theta_s$; {\sc Abs\textsubscript{pp-ub}} is the upper bound for large-scale alignment) or using a mixture of pseudo-parallel and synthetic data ({\sc Abs\textsubscript{pp+syn-N}}, using the {\sc Abs\textsubscript{pp-0.63}} dataset and backtranslated data from the top $N$ beam hypotheses). We always use the {\sc Lead} extractor.}
    \label{tab:synthetic}

\end{table}

\paragraph{Implementation details.}

$P_\text{PM}$ and $P_\text{BT}$ are both implemented as bidirectional LSTM encoder-decoder models with 256 hidden units, embedding dimension 128, and an attention mechanism \cite{bahdanau2014neural}. We pick this model size to be comparable to recent work \cite{see2017get,chen2018fast}. We set the vocabulary size to 50k and train both models until convergence with Adam \cite{kingma2014adam}; $P_\text{PM}$ uses beam search with a beam of 5 during testing. 

\paragraph{}

Because both of our extractor variants are unsupervised, we directly apply them to the articles to select salient sentences. We always set the number $K$ of sentences to be extracted to the average number of summary sentences in the target dataset, $K=4$ for the CNN/DailyMail dataset, and $K=25$ for the scientific article dataset. 

\paragraph{Evaluation details.} We evaluate our systems using the ROUGE-1/2/L F1 metrics \cite{lin2004rouge} and METEOR \cite{banerjee2005meteor}, which is often used in machine translation. 

\section{Experiments on CNN/DailyMail}

We use the CNN/DailyMail (CNN/DM) dataset \cite{hermann2015teaching} consisting of pairs of news articles from CNN and Daily Mail, along with summaries in the form of bullet points. We choose this dataset because it allows us to compare our approach to existing fully supervised methods and to measure the gap between unsupervised and supervised summarization. We follow the preprocessing pipeline of \cite{chen2018fast}, splitting the dataset into 287k/11k/11k pairs for training/validation/testing. Note that our method relies only on the bullet-point summaries from this training set.

\paragraph{Obtaining synthetic data.}

To obtain training data for our sentence abstractor $P_\text{PM}$, we follow the procedure from Section \ref{sec:abstractor}. We align all summaries from the CNN/DM training set to 8.5M news articles from the Gigaword dataset \cite{napoles2012annotated}, which \textit{contains no articles from CNN or Daily Mail}. After alignment\footnote{We follow the set-up from \cite{nikolov2018large} using the Sent2Vec embedding method \cite{pagliardini2017unsupervised} for computing document/sentence embeddings. We use hyperparameters $\theta_d=0.5$ and $\theta_s=\{0.60, 0.63, 0.67\}$.}, we obtain 1.2M pseudo-parallel setnence pairs that we use to train our backtranslation model $P_\text{BT}$. Using $P_\text{BT}$, we synthesize $J=5$ article sentences for each of the ~1M summary sentences by picking the top 5 beam hypotheses. Our best sentence paraphrasing dataset used to train our final abstractor $P_\text{PM}$ contains 6.7 million sentence pairs, 18\% of which are pseudo-parallel pairs and 82\% are backtranslated pairs. 

\subsection{Results on CNN/DailyMail}\label{sec:automatic-eval} 

\paragraph{Baselines.} We compare our models with several supervised and unsupervised baselines. {\sc LSTM} is a standard bidirectional LSTM model trained to directly generate the CNN/DM summaries from the full CNN/DM articles. {\sc Ext-Abs} is a hierarchical model consisting of a supervised LSTM extractor and separate abstractor \cite{chen2018fast}, both of which are individually trained on the CNN/DM dataset by aligning summary to article sentences. Our work best resembles {\sc Ext-Abs} except that we do not rely on any parallel data. {\sc Ext-Abs-RL} is a state-of-the-art summarization system that extends {\sc Ext-Abs} by jointly tuning the two supervised components using reinforcement learning. We additionally report the performance of our unsupervised extractive baselines, {\sc Lead}, and {\sc LexRank}, as well as the result of an oracle ({\sc Oracle}) which computes an upper bound for extractive summarization by aligning the ground truth summary sentences to their original articles using ROUGE-1. 

\paragraph{Automatic evaluation.}

Our best abstractive models trained on non-parallel data ({\sc Lead + Abs\textsubscript{pp+syn-5}} and {\sc LexRank + Abs\textsubscript{pp+syn-5}} in Table \ref{tab:results-cnndm}) performed worse than the baselines trained on parallel data. However, the results are promising: for example, the ROUGE-L gap between our {\sc Lead} model and the supervised {\sc LSTM} model is only $1.8$. When comparing against the {\sc Ext-Abs} and {\sc Ext-Abs-RL} models, which perform supervised sentence extraction followed by supervised sentence abstraction, the gap is larger. Furthermore, we observe that, on this dataset, applying our abstractors to the {\sc Lead} and {\sc LexRank} extractive baselines leads to a decrease in ROUGE. This could be due to the much shorter length of our abstractive summaries in comparison to the length produced by other systems, indicating that our systems potentially summarize much more aggressively. 

\paragraph{Model analysis.} 

In Table \ref{tab:synthetic}, we compare the effect of training our abstractor on pseudo-parallel datasets of different sizes ({\sc Abs\textsubscript{pp-*}}) as well as on a mixture of pseudo-parallel and backtranslated data ({\sc Abs\textsubscript{pp+syn-*}}). For reference, we also include results from aligning the original dataset of CNN/DM articles and summaries directly. We construct a \textit{parallel} dataset ({\sc Abs\textsubscript{par}}) of sentence pairs by aligning the original CNN/DM document pairs using ROUGE-1; as well as a pseudo-parallel dataset ({\sc Abs\textsubscript{pp-ub}}) by applying the large-scale alignment method to the CNN/DM documents, (without using the document labels directly). The performance difference between {\sc Abs\textsubscript{par}} and {\sc Abs\textsubscript{pp-ub}} provides an estimate of the performance loss due to pseudo alignment. 

\begin{table*}
\RawFloats
\centering
\small 
\begin{tabular}{c|c|c|c|c|c}
\hlineB{2}
\textbf{Approach} & \textbf{R-1} & \textbf{R-2} & \textbf{R-L} & \textbf{MET} & \textbf{\#} \\ \hlineB{2}
{\sc Oracle} & 43.79 & 14.27 & 41.1 & 20.05 & 969 \\ \hlineB{2}
\multicolumn{6}{c}{\textbf{Unsupervised extractive baselines}} \\ \hlineB{2}
{\sc Lead} & 41.2 & 11.85 & 38.87 & 17.11 & 688 \\ 
{\sc LexRank} & 38.78 & 10.31 & 36.57 & 16.66 & 800 \\ \hlineB{2}
\multicolumn{6}{c}{\textbf{Abstractive summarization without parallel data (this work)}} \\ \hlineB{2}
{\sc Lead + Abs\textsubscript{pp+syn-1}} & 42.47 & 12.11 & 40.25 & 15.78 & 529 \\ 
{\sc LexRank + Abs\textsubscript{pp+syn-1}} & 41.04 & 10.94 & 38.9 & 15.38 & 562 \\ \hlineB{2}
\end{tabular}
\protect\caption[position=bottom]{Metric results on the Scientific summarization test set. \textbf{R-1/2/L} are the \textbf{ROUGE-1/2/L} F1 scores; \textbf{MET} is \textbf{METEOR}, while \textbf{\#} is the average number of tokens in the summaries.}
\label{tab:results-science}
\end{table*}
\begin{table}
\small
\centering
\caption{Datasets used to extract pseudo-parallel monolingual sentence pairs in our style transfer experiments.}
\label{table:datasets}
\begin{tabular}{c|c|c|c}
\hlineB{2}
\textbf{Dataset} & \textbf{Documents} & \textbf{Tok. per sent.} & \textbf{Sent. per doc.} \\ \hlineB{2}
PubMed & 1.5M & 24 $\pm$ 14 & 180 $\pm$ 98 \\ 
MEDLINE & 16.8M & 26 $\pm$ 13 & 7 $\pm$ 4 \\ 
EurekAlert & 358K & 29 $\pm$ 15 & 25 $\pm$ 13 \\
Wikipedia & 5.5M & 25 $\pm$ 16 & 17 $\pm$ 32 \\
\hlineB{2}
\end{tabular}
\end{table}

Our best pseudo-parallel abstractor performs poorly in comparison to the parallel abstractor. Adding additional synthetic data is helpful but insufficient to compensate for the performance gap. Furthermore, we observe a diminishing improvement from adding synthetic pairs. By contrast, the large-scale alignment method constructs a pseudo-parallel upper bound that almost perfectly matches the parallel dataset, indicating that potentially the main bottleneck in our system is the domain difference between the articles in Gigaword and the CNN/DM. 

\paragraph{Example summaries.}

In Table \ref{tab:examples}, we also provide example summaries produced on the CNN/DM dataset. Our final model trained on additional backtranslated data produced much more relevant and coherent sentences than the model trained on pseudo-parallel data only. Despite having seen no parallel examples, the system is capable of generating fluent, abstractive sentences. However, in comparison to the abstractor trained on parallel data, there is still room for further improvement. 

\section{Experiments on Scientific Articles}\label{sec:science}

The rate of scientific publications grows exponentially \cite{hunter2006biomedical}, calling for efficient automatic summarization tools \cite{science-summ}. An important frontier in scientific text summarization is generating summaries that not only synthesize an article but make it also more accessible to non-specialists \cite{vadapalli2018sci,vadapalli2018science,tatalovic2018ai}. A major challenge towards achieving this goal is the lack of parallel datasets of articles and high-quality summaries. 

\paragraph{}

We introduce the novel task of automatically generating a press release for a scientific article. Although there are already large repositories of scientific articles and press releases, two major obstacles are preventing manual alignment across these resources: 
\begin{enumerate}
    \item Because press releases address very different audiences, they are often written and published separately from the article. Furthermore, there is often no metadata present in the text (such as a digital object identifier (DOI)) that could be used to link a press release to its original scientific article. 
    \item Even when there is metadata that can be utilized, the full text of the original article may not be accessible, either because it is published in a closed access journal under a restrictive license, or because the full text of the article is not available in an easily parsable format (e.g., only a PDF is available).  
\end{enumerate}

These practical challenges call for alternative summarization approaches. The summarization method described in this paper circumvents these problems because it exploits the large existing collections of papers and press releases that exist within open repositories. 

\paragraph{Datasets used for alignment.}

To extract pseudo-parallel sentence pairs for paraphrasing, we rely on four collections of documents. We combine two datasets of scientific articles: \texttt{PubMed}\footnote{\small{\url{www.ncbi.nlm.nih.gov/pmc/tools/openftlist}}}, which contains the full text of $1.5M$ open access papers, and \texttt{Medline}\footnote{\url{www.nlm.nih.gov/bsd/medline.html}}, which contains over $17M$ scientific abstracts. We obtain press releases through scraping $350k$ articles from Eurekalert\footnote{\url{https://www.eurekalert.org/}}, an aggregator that covers many scientific disciplines. As an additional resource that contains scientific texts written for a more general audience, we use all articles on Wikipedia, without applying any filtering. Table \ref{table:datasets} contains an overview of these datasets. 

\paragraph{Evaluation dataset.}  

To create a parallel testing dataset, we use regular expressions to detect digital object identifier (DOI) mentions within the press releases. Given a DOI, we query for the full text of a paper using the Elsevier ScienceDirect API\footnote{\url{https://dev.elsevier.com/sd_apis.html}}. Using this approach, we were able to compose 6821 parallel pairs of the full text of a paper and its press release (which amounts to only $2\%$ of all press releases available). We use these 6821 pairs as our testing set in our experiments, and exclude them from the alignment procedure. 

\paragraph{Extracting pseudo-parallel data.}

To obtain a pseudo-parallel dataset for scientific sentence paraphrasing, we aligned the scientific papers to the press releases. After alignment\footnote{We use a document similarity threshold $\theta_d=0.6$ and a sentence similarity threshold $\theta_s=0.74$.}, we extracted $80k$ pseudo-parallel pairs in total. We additionally aligned \texttt{PubMed} and \texttt{Medline} to all articles on \texttt{Wikipedia}\footnote{Using $\theta_d=0.6$ and  $\theta_s=0.78$.}, obtaining out-of-domain pairs for this task. We merged all pairs into a single dataset, consisting of $370k$ sentence pairs. We used these pairs to train a backtranslation model $P_\text{BT}$ from which we synthesized 1 article sentence for each sentence in the press release dataset\footnote{We synthesize only a single sentence because the press release dataset contains many more sentences than CNN/DailyMail.}. Our final sentence paraphrasing dataset used to train the abstractor $P_\text{PM}$ contains $8.6M$ pairs. 

\subsection{Results}

\begin{table}
    \centering
		\small 
		\begin{tabular}{p{7.5cm}}
    \hlineB{2} 
    {\sc Lead: } global reference models , such as preliminary reference earth model ( prem ; dziewonski and anderson , 1981 ) , iasp91 ( kennett and engdahl , 1991 ) and ak135 ( kennett et al. , 1995 ) were created with different data sets , techniques and assumptions , leading to differences in some regions of earth . \\ \hline
   {\sc Lead + Abs\textsubscript{pp+syn-1}: } the study was created with different data sets , techniques and assumptions , leading to differences in some regions of earth . \\ \hlineB{2}
   {\sc Lead: } astroglia from the postnatal cerebral cortex can be reprogrammed in vitro to generate neurons following forced expression of neurogenic transcription factors , thus opening new avenues towards a potential use of endogenous astroglia for brain repair . \\ \hline
   {\sc Lead + Abs\textsubscript{pp+syn-1}:} the brain ' s neural network has been reprogrammed to generate neurons from the postnatal cerebral cortex . \\ \hlineB{2}
   {\sc Lead: } sequenced the angiotensinogen gene and found that multiple rare variants contribute to variation in angiotensinogen levels ; interestingly , most of these variants sit on the same haplotype background created by three common snps.26 fourth , three common tag snps encompassing mc1r ( mim 155555 ) were independently associated with melanoma in a recent gwas.27 however , resequencing of the candidate gene , mc1r , indicates that these signals can be completely explained by the combined effects of several rare nonsynonymous mutations , suggesting that ignoring rare variants can lead to incorrect inferences on the potential role of candidate genes carrying common snps identified by gwas ( f. demenais at al . \\ \hline
   {\sc Lead + Abs\textsubscript{pp+syn-1}:} the researchers found that these variants can lead to incorrect inferences on the potential role of candidate genes carrying common snps . \\ \hlineB{2}
    \end{tabular}
	\caption{Example sentence paraphrases produced on the scientific dataset (using the {\sc Lead} extractor).}
	\label{tab:science-examples}
\end{table}

\paragraph{Baselines.} 

Since there are no existing parallel datasets of scientific articles and press releases, here we do not report results using supervised methods. We compare the performance of our models ({\sc Lead + Abs\textsubscript{pp+syn-1}} and {\sc LexRank + Abs\textsubscript{pp+syn-1}}) to the two unsupervised extractive baselines, {\sc Lead} and {\sc LexRank} respectively, as well as to the result of the {\sc Oracle} (same as in Section \ref{sec:automatic-eval})  that yields an upper bound for extractive summarization. 

\paragraph{Model analysis.} 

Our results are in Table \ref{tab:results-science}. Both of our abstractive models outperformed their respective extractive baselines, indicating that our abstractors are beneficial for this task and have learned useful sentence transformations. The summaries are much shorter after abstraction, indicating that the paraphrased summaries are meaningfully compressed. Surprisingly, the gap between the baselines and the oracle is small, indicating that abstractive summarization is essential for achieving good performance on this task.

\paragraph{Example summaries.} 

We find that the model trained on backtranslated data is often conservative choosing to leave many input sentences almost unchanged. However, as shown by examples in Table \ref{tab:science-examples}, the model learned useful transformations by compressing long sentences and utilizing vocabulary adapted to the language of press releases (e.g. "the researchers" in the last example). The sentence compressions can sometimes seem too drastic when looked at in isolation (e.g., from "astroglia from the postnatal cerebral cortex" to "the brain's neural network" in the second example). 

\section{Conclusion}

We developed an abstractive summarization system that does not rely on parallel resources, but can instead be trained using example summaries and a large collection of non-matching articles, making it particularly relevant to low-resource domains and languages. On the CNN/DailyMail benchmark, our system performed competitively to a fully supervised LSTM baseline trained on the document level. We also achieved promising results on the novel task of automatically generating a press release for a scientific journal article. 

\paragraph{}

Future work will focus on developing novel unsupervised extractors, on decreasing the gap between abstractors trained on parallel and non-parallel data, as well as on developing methods for combining the abstractor and extractor into a single system. 

\section*{Acknowledgments}

We acknowledge support from the Swiss National Science Foundation (grant $31003A\_156976$) and thank the anonymous reviewers for their useful comments.  

\section{Bibliographical References}
\label{main:ref}

\bibliographystyle{lrec}


\end{document}